\documentclass[conference,a4paper]{IEEEtran}
\IEEEoverridecommandlockouts

\usepackage{cite}
\usepackage{amsmath,amssymb,amsfonts}
\usepackage{algorithmic}
\usepackage{graphicx}
\usepackage{textcomp}
\usepackage{xcolor}
\usepackage{booktabs} 
\usepackage{url} 
\usepackage{hyperref} 

\def\BibTeX{{\rm B\kern-.05em{\sc i\kern-.025em b}\kern-.08em
    T\kern-.1667em\lower.7ex\hbox{E}\kern-.125emX}}

\begin{document}

\title{KyrgyzBERT: A Compact, Efficient Language Model for Kyrgyz NLP}

\author{
    \IEEEauthorblockN{Adilet Metinov\IEEEauthorrefmark{1},
                      Gulida M. Kudakeeva\IEEEauthorrefmark{2},
                      Gulnara D. Kabaeva\IEEEauthorrefmark{3}}
    \IEEEauthorblockA{\IEEEauthorrefmark{1}\IEEEauthorrefmark{2}\IEEEauthorrefmark{3}%
    Institute of Information Technology,\\
    Kyrgyz State Technical University named after I. Razzakov, Bishkek, Kyrgyzstan}
    \IEEEauthorblockA{\IEEEauthorrefmark{1}\texttt{metinovab@kstu.kg},\;
                      \IEEEauthorrefmark{2}\texttt{kgm@kstu.kg},\;
                      \IEEEauthorrefmark{3}\texttt{kabaevagd9@kstu.kg}}
}

\maketitle

\begin{abstract}
The advancement of Natural Language Processing (NLP) has been largely driven by large-scale transformer models, yet these breakthroughs have predominantly benefited high-resource languages. Low-resource languages like Kyrgyz lack the foundational models necessary for robust NLP application development. To address this gap, we introduce \texttt{KyrgyzBert}, the first publicly available, monolingual BERT-based language model pre-trained for the Kyrgyz language. KyrgyzBERT is a compact and efficient 35.9M parameter model developed with a custom tokenizer to optimally handle the language's unique morphological characteristics. To demonstrate its effectiveness, we construct a new, high-quality sentiment analysis benchmark, \texttt{kyrgyz-sst2}, by translating the Stanford Sentiment Treebank and manually annotating the entire test set. Our experiments show that our finetuned model, \texttt{kyrgyzbert\_sst2}, achieves an F1-score of 0.8280, a result highly competitive with a finetuned mBERT model five times its size. We publicly release all artifacts from this study, providing a comprehensive suite of tools to accelerate NLP research for the Kyrgyz language.
\end{abstract}

\begin{IEEEkeywords}
Natural Language Processing, Kyrgyz Language, Language Model, BERT, Low-resource Languages, Sentiment Analysis, Computational Efficiency
\end{IEEEkeywords}

\section{Introduction}
Recent years have witnessed a paradigm shift in Natural Language Processing (NLP), spurred by the advent of large-scale, transformer-based language models such as BERT \cite{devlin2019bert}. These models, pre-trained on vast amounts of text data, have established new state-of-the-art results across a wide array of tasks. However, this progress has created a significant resource disparity; languages like English have a rich ecosystem of models and datasets, while low-resource languages like Kyrgyz remain critically underserved.

The Kyrgyz language, a member of the Turkic language family, lacks the foundational computational resources necessary to leverage modern NLP techniques. This technological gap hinders both academic research and the development of practical applications for the Kyrgyz-speaking community. While multilingual models like mBERT \cite{devlin2019bert} and XLM-RoBERTa \cite{conneau2020unsupervised} offer a potential starting point, they are often sub-optimal. Recent studies have shown that general-purpose multilingual models can underperform compared to classical machine learning methods on Kyrgyz tasks, highlighting the need for specialized resources \cite{benli2024using}. These models must balance representational capacity across over 100 languages and may not fully capture the rich, agglutinative morphology of Kyrgyz.

To address this critical gap, we present two primary contributions. First, we introduce \textbf{KyrgyzBERT}, the first publicly available monolingual BERT model specifically pre-trained for the Kyrgyz language. To create this model, we developed a custom WordPiece tokenizer from scratch and pre-trained a compact, 35.9M parameter model on a large corpus of Kyrgyz text. Our goal is to provide a computationally efficient and linguistically tailored foundation for future Kyrgyz NLP research.

Our second contribution is a rigorous validation of KyrgyzBERT's capabilities. To this end, we constructed a new, high-quality sentiment analysis benchmark dataset based on the Stanford Sentiment Treebank (SST-2) \cite{socher2013recursive}. Crucially, to ensure the reliability of our evaluation, the entire test set was manually created and annotated by a native speaker after translation. We use this benchmark to finetune and evaluate KyrgyzBERT against strong multilingual baselines. Our experiments demonstrate that KyrgyzBERT achieves performance highly competitive with mBERT, despite being five times smaller.

All artifacts from this study—the pre-trained \texttt{KyrgyzBert}, the custom \texttt{bert-kyrgyz-tokenizer}, the finetuned sentiment models, and the \texttt{kyrgyz-sst2} benchmark dataset—are made publicly available\footnote{All artifacts from this study are publicly available at: \url{https://huggingface.co/metinovadilet}} to foster and accelerate the development of NLP technologies for the Kyrgyz language.

\section{Related Work}
Research in NLP for Turkic languages has grown, but resources for Kyrgyz remain scarce. While models for Turkish, such as BERTurk \cite{schweter2020berturk}, and for Arabic, like AraBERT \cite{antoun2020arabert}, have demonstrated the value of monolingual transformers, Kyrgyz has largely relied on multilingual models or classical machine learning approaches.

A recent and highly relevant study by Benli and Sharshembaev \cite{benli2024using} conducted a comparative analysis of sentiment analysis for Kyrgyz. They used a dataset of 500 IMDB reviews translated into Kyrgyz and tested a range of models, including Logistic Regression (LR), Random Forest (RF), and deep learning models like LSTM and RNN. Their key finding was that a traditional LR model achieved the highest accuracy (83\%), outperforming a multilingual BERT model which only scored 60\%. This result highlighted the sub-optimal performance of general multilingual models on Kyrgyz when not specifically adapted for the task. Crucially, the authors concluded that future progress in Kyrgyz NLP requires the development of a \textbf{Kyrgyz-specific BERT model} and larger, higher-quality datasets. Our work directly addresses this call by introducing KyrgyzBERT and a new, manually-verified benchmark dataset.

\section{Methodology}
Our methodology is centered around two core efforts: the creation of the foundational KyrgyzBERT model and the construction of a high-quality benchmark for its evaluation.

\subsection{KyrgyzBERT Pre-training}
The cornerstone of our work is \texttt{metinovadilet/KyrgyzBert}, a monolingual language model designed specifically for Kyrgyz.

\subsubsection{Tokenizer}
A standard multilingual vocabulary is ill-suited for the agglutinative nature of Kyrgyz. Therefore, we first trained a new WordPiece tokenizer from scratch, \texttt{metinovadilet/bert-kyrgyz-tokenizer}, on a large Kyrgyz text corpus, ensuring optimal segmentation of Kyrgyz words into meaningful sub-word units.

\subsubsection{Model Architecture and Training}
`KyrgyzBert` is a BERT-based model with a compact architecture, making it efficient for training and inference. The key specifications are detailed in Table \ref{tab:kyrgyzbert_arch}. The model was pre-trained from scratch on a private corpus of over 1.5 million Kyrgyz sentences. Training was performed on a single NVIDIA RTX 3090 GPU using the Masked Language Modeling (MLM) objective.

\begin{table}[htbp]
\caption{KyrgyzBERT Model Architecture}
\centering
\begin{tabular}{lc}
\toprule
\textbf{Parameter} & \textbf{Value} \\
\midrule
Vocabulary Size & 30,522 \\
Hidden Size & 512 \\
Attention Heads & 8 \\
Number of Layers & 6 \\
Total Parameters & 35.9M \\
\bottomrule
\end{tabular}
\label{tab:kyrgyzbert_arch}
\end{table}

\subsection{Sentiment Analysis Benchmark}
To validate KyrgyzBERT, we constructed a new sentiment analysis benchmark, which we name and release as \texttt{metinovadilet/kyrgyz-sst2}.

\subsubsection{Dataset Creation}
We used the Stanford Sentiment Treebank (SST-2) as our source. The English sentences from the original train and validation splits were translated into Kyrgyz using a state-of-the-art Neural Machine Translation (NMT) system, with their original labels carried over.

\subsubsection{Gold-Standard Test Set Curation}
To create a reliable evaluation set, the 1,821 **unlabeled** sentences from the original SST-2 test split were first translated into Kyrgyz using an NMT system. Subsequently, each of these translated Kyrgyz sentences was **manually annotated for sentiment by a native Kyrgyz speaker.** This crucial step ensures that our test set's labels are not merely carried over from an English source but are ground-truth assessments of the sentiment expressed in the final Kyrgyz text, creating a 'gold-standard' for evaluation.

\subsection{Experimental Setup}
We conducted a comparative sentiment analysis experiment. The finetuned models are the focus of our analysis, with the base models serving as a zero-shot baseline.
All models were finetuned for 3 epochs with a learning rate of 2e-5 where applicable. We evaluated performance using the weighted F1-score and Accuracy.

\section{Results and Discussion}
\begin{figure}[htbp]
\centering 
\includegraphics[width=\columnwidth]{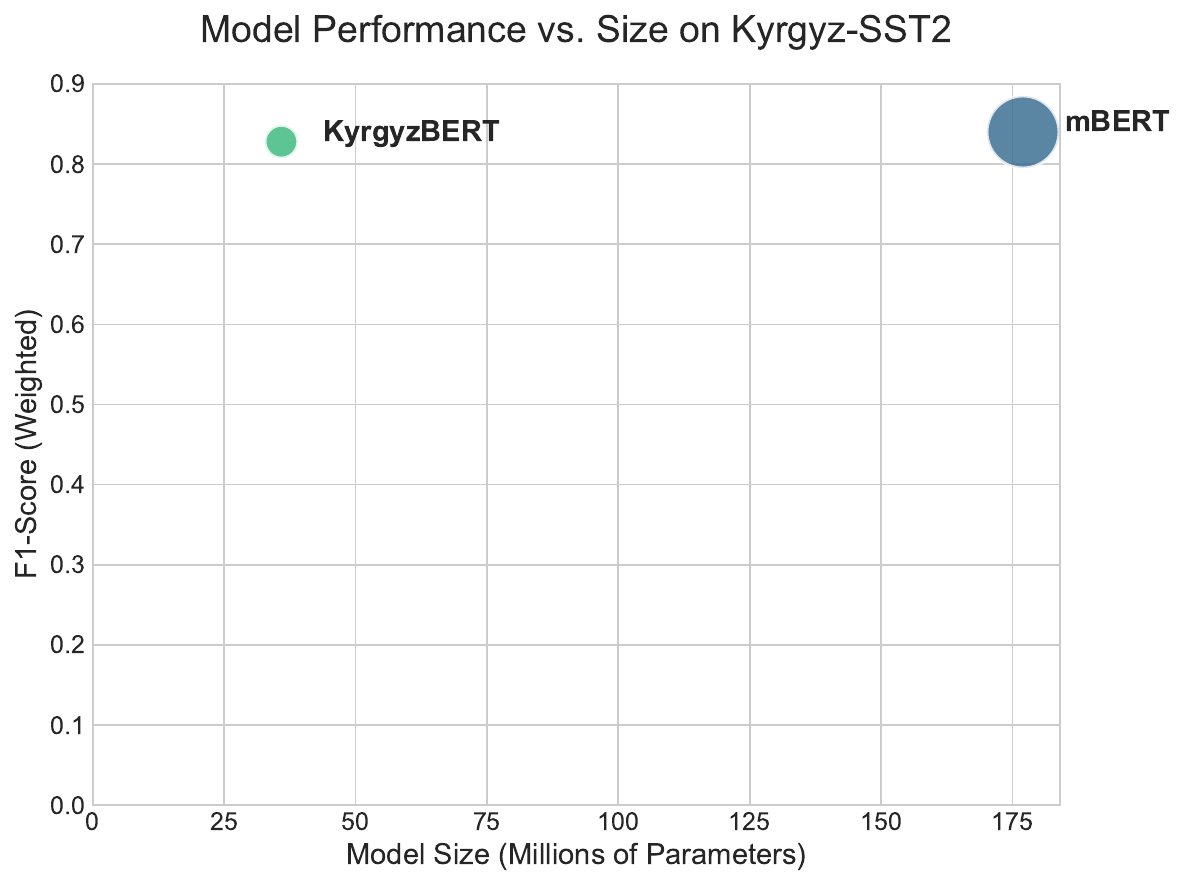}
\caption{Model performance (F1-Score) vs. size (millions of parameters). KyrgyzBERT achieves performance competitive with mBERT while being 5x smaller, demonstrating superior computational efficiency.}
\label{fig:perf_vs_size}
\end{figure}

The comprehensive results of our evaluation are presented in Table \ref{tab:results}, with the performance-to-size trade-off for the finetuned models visualized in Fig. \ref{fig:perf_vs_size}.

\begin{table}[htbp]
\caption{Sentiment Analysis Benchmark Results}
\centering
\begin{tabular}{lcc}
\toprule
\textbf{Model} & \textbf{F1-score (Weighted)} & \textbf{Size (M Params)} \\
\midrule
\multicolumn{3}{l}{\textit{Finetuned Models}} \\
\textbf{KyrgyzBERT} & 0.8280 & \textbf{35.9} \\
mBERT & \textbf{0.8401} & 177.0 \\
\midrule
\multicolumn{3}{l}{\textit{Base Models (Zero-Shot)}} \\
XLM-RoBERTa & 0.3221 & 270.0 \\
mBERT & 0.3509 & 177.0 \\
\bottomrule
\end{tabular}
\label{tab:results}
\end{table}

The results clearly indicate that task-specific finetuning is essential, as the base models perform poorly, with F1-scores near the level of a random classifier (Table \ref{tab:results}). Our primary finding is that finetuned `KyrgyzBert` achieves a strong F1-score of 0.8280, confirming that a compact, monolingual model is highly effective for downstream tasks in Kyrgyz.

As visualized in Fig. \ref{fig:perf_vs_size}, the most compelling result is the trade-off between performance and computational efficiency. `KyrgyzBert` is approximately five times smaller than mBERT, yet the performance gap is only 1.2\%. This exceptional efficiency is a critical advantage for a low-resource language community, as it makes advanced NLP models more accessible by lowering the barrier for training, finetuning, and deployment on consumer-grade hardware. Interestingly, our finetuned mBERT achieves a high F1-score of 0.84, in contrast to the 60\% accuracy reported by Benli and Sharshembaev \cite{benli2024using} for a non-finetuned mBERT, underscoring the critical importance of task-specific finetuning and high-quality benchmark data.

The immediate adoption of `KyrgyzBert` by the community further underscores its value. Since its release on the Hugging Face Hub, the model has been downloaded over 300 times, indicating a clear and active demand for such foundational resources.

\section{Conclusion}
In this paper, we introduced `KyrgyzBert`, a compact and efficient 35.9M parameter language model, and its accompanying tokenizer, as new foundational resources for the Kyrgyz language. By evaluating it on a newly created and manually annotated sentiment analysis benchmark, we have demonstrated its strong performance, achieving results competitive with models five times its size.

Our work makes a significant contribution by providing the first suite of open-source, monolingual tools specifically for Kyrgyz NLP. All developed resources—including the base \texttt{metinovadilet/KyrgyzBert}, the finetuned sentiment models \texttt{metinovadilet/kyrgyzbert\_sst2} and \texttt{metinovadilet/mbert-kyrgyz-sst2-finetuned}, and the \texttt{metinovadilet/kyrgyz-sst2} dataset—are publicly available to foster a new wave of research and application development. Future work should focus on evaluating `KyrgyzBert` across more diverse tasks and domains, and pre-training larger versions as more data becomes available.

\section*{Acknowledgment}
The authors would like to thank the colleagues and administration of the Institute of Information Technology, Kyrgyz State Technical University, for their support in providing computational resources and an environment that enabled this research.

\bibliographystyle{IEEEtran}
\bibliography{references}

@article{devlin2019bert,
  title={BERT: Pre-training of Deep Bidirectional Transformers for Language Understanding},
  author={Devlin, Jacob and Chang, Ming-Wei and Lee, Kenton and Toutanova, Kristina},
  journal={arXiv preprint arXiv:1810.04805},
  year={2018}
}

@article{conneau2020unsupervised,
  title={Unsupervised cross-lingual representation learning at scale},
  author={Conneau, Alexis and Khandelwal, Kartik and Goyal, Naman and Chaudhary, Vishrav and Wenzek, Guillaume and Guzm{\'a}n, Francisco and Grave, Edouard and Ott, Myle and Zettlemoyer, Luke and Stoyanov, Veselin},
  journal={arXiv preprint arXiv:1911.02116},
  year={2019}
}

@inproceedings{socher2013recursive,
  title={Recursive deep models for semantic compositionality over a sentiment treebank},
  author={Socher, Richard and Perelygin, Alex and Wu, Jean and Chuang, Jason and Manning, Christopher D and Ng, Andrew Y and Potts, Christopher},
  booktitle={Proceedings of the 2013 conference on empirical methods in natural language processing},
  pages={1631--1642},
  year={2013}
}

@article{benli2024using,
  title={{Using Machine Learning Algorithms for Kyrgyz Sentiment Analysis}},
  author={Benli, {\.I}brahim and Sharshembaev, Bakyt},
  journal={World Journal of Advanced Research and Reviews},
  volume={23},
  number={03},
  pages={554--561},
  year={2024},
  doi={10.30574/wjarr.2024.23.3.2681}
}

@article{schweter2020berturk,
  title={{BERT}urk: A Turkish {BERT} model},
  author={Schweter, Stefan},
  journal={arXiv preprint arXiv:2005.00570},
  year={2020}
}

@inproceedings{antoun2020arabert,
  title={{A}ra{BERT}: Transformer-based Model for Arabic Language Understanding},
  author={Antoun, Wissam and Baly, Fady and Hajj, Hazem},
  booktitle={Proceedings of the 4th Workshop on Open-Source Arabic Corpora and Processing Tools, with a Shared Task on Offensive Language Detection},
  pages={9--15},
  year={2020}
}

\end{document}